\documentclass[letterpaper]{article}
\usepackage{aaai}
\usepackage{times}
\usepackage{helvet}
\usepackage{courier}
\usepackage{nicefrac}
\usepackage{xr}

\usepackage{times}
\usepackage{color}
\usepackage{latexsym}

\usepackage{graphicx}
\usepackage[centertags,fleqn]{amsmath}
\usepackage{amssymb,amsfonts,xspace}
\usepackage{booktabs}
\usepackage{multirow}
\title{The Singularity May Never Be Near}

\author{Toby Walsh\\ University of New South Wales
and Data61 (formerly NICTA)\\ Sydney, Australia}

\begin{document}
\maketitle

\section{Introduction}

There is both much optimism and pessimism around artificial
intelligence (AI) today. 
The optimists are investing millions of dollars, 
and even in some cases billions of dollars into AI. 
The pessimists, on the other hand, predict
that AI will end many things: jobs, 
warfare, and even the human race. 
Both the optimists and
the pessimists often appeal to the idea of a technological
singularity, a point in time where machine intelligence
starts to run away, and a new, more intelligent
``species'' starts to inhabit the earth. 
If the optimists are right, this will be
a moment that fundamentally changes our 
economy and our society. If the pessimists
are right, this will be a moment that 
also fundamentally changes our economy
and our society. It is therefore very worthwhile
spending some time deciding if either of 
them might be right. 

\section{The History of the Singularity}

The idea of a technological singularity can be traced back to 
a number of different thinkers. 
Following John von Neumann's death in 1957, Stanislaw Ulam wrote:
\begin{quote}
``{\em One conversation [with John von Neumann]
centered on the ever accelerating progress of technology and changes 
in the mode of human life, which gives the appearance of approaching 
some essential singularity in the history of the race beyond which 
human affairs, as we know them, could not continue.}''  \cite{ulam}
\end{quote}
I.J. Good made a more specific prediction in 1965, 
calling it an ``intelligence explosion'' rather than a
``singularity'':
\begin{quote}
``{\em Let an ultraintelligent machine be defined as a machine that can far surpass all the
intellectual activities of any man however clever. Since the design of machines is one
of these intellectual activities,  an ultraintelligent machine could design even better
machines; there would then unquestionably be an “intelligence explosion”, and the
intelligence of man would be left far behind. Thus the first ultraintelligent machine is
the last invention that man need ever make.'' }\cite{good}
\end{quote}
Many credit the technological singularity
to the computer scientist, and science fiction author Vernor Vinge
who predicted:
\begin{quote}
``{\em    Within thirty years, we will have the technological means to create superhuman intelligence. Shortly after, the human era will be ended.}'' \cite{vinge}
\end{quote}
More recently, the idea of a technological singularity
has been popularized by Ray Kurzweil \cite{kurzweil} as well as 
others. 
Based on current trends, Kurzweil predicts the technological
singularity will happen around 2045. 
For the purposes of this
article, I suppose that the technological singularity is
the point in time at which we build a machine of sufficient
intelligence that is able to
redesign itself to improve its intelligence, and at which
its intelligence starts to grow exponentially fast, 
quickly exceeding human intelligence by orders of magnitude.

I start with two mathematical quibbles. The first quibble is
that the technological
singularity is not a mathematical singularity. The
function $\frac{1}{1-t}$ has a mathematical singularity at $t=1$. 
This function demonstrates hyperbolic growth. As $t$ approaches
1, its derivative ceases to be finite and well defined. 
Many proponents of a technological singularity argue
only for exponential growth. For exampe,
the function $2^t$ demonstrates exponential
growth. Such an exponential function approaches infinity more slowly, and 
has a finite derivative that is always well defined. The second
quibble is that the idea of exponential growth in intelligence depends 
entirely on the scale used to measure intelligence.
If we measure intelligence in logspace, exponential growth is merely
linear. I will not tackle here head on what we mean
by measuring the intelligence of machines (or of humans).
I will simply suppose there is such a property as intelligence, that
it can be measured and compared, and that
the technological singularity is when this measure
increases exponentially fast in an appropriate and reasonable
scale. 

The possibility of a technological singularity has driven
several commentators to issue dire
predictions about the possible impact of artificial intelligence
on the human race.
For instance, in December 2014, Stephen Hawking told the BBC:
\begin{quote}
``{\em  The development of full artificial intelligence could spell the end of the human race. \ldots It would take off on its own, and re-design itself at an ever increasing rate. Humans, who are limited by slow biological evolution, couldn't compete, and would be superseded.
}''
\end{quote}
Several other well known figures including
Bill Gates, Elon Musk and Steve Wozniak 
have subsequently issued similar warnings. 
Nick Bostrom has
predicted a technological singularity,
and argued that this poses an existential threat
to the human race \cite{bostrom}. 
In this article, I will explore arguments as to
why a technological singularity might not be 
close.

\section{Some arguments against the Singularity}

The idea of a technological singularity has 
received more debate outside 
the mainstream AI community than within it. 
In part, this may be because many of the proponents for such an event
have come from outside this community. 
The technological singularity also has become associated
with some somewhat challenging 
ideas like life extension and transhumanism. This
is unfortunate as it has distracted debate from a
fundamental and important issue: will we able to develop machines
that at some point are able to improve their intelligence exponentially fast and that
quickly far exceed our own human intelligence? 
This might not seem a particularly wild idea.
The field of computing has profited considerably from exponential trends.
Moore's Law has predicted with reasonable accuracy
that the number of transistors on an integrated circuit
(and hence the amount of memory in a chip) will double every two years
since 1975. 
And Koomey’s law has accurately predicted that the number of 
computations per joule of energy dissipated will double every 19 months
since the 1950s.
Is it unreasonable to suppose AI will also at 
some point witness exponential growth? 

The thesis put forwards here is that there are several strong
arguments against the possibility of a technological
singularity. Let me be precise. I am not 
predicting that AI will fail to achieve super-human
intelligence. Like many of my colleagues working in 
AI, I predict we are just 30 or 40 years away from this 
event. However, I am suggesting that there 
will not be the run away exponential growth predicted by 
some. I will put forwards multiple arguments why a technological 
singularity is improbable. 

These are not the only arguments against 
a technological singularity. We can, for instance,
also inherit all the arguments raised against 
artificial intelligence itself. Hence, there are also the nine common objections
considered by Alan Turing in his seminal Mind
paper \cite{turing2} like machines not
being conscious, or not being creative. 
I focus here though on arguments which go to
the idea of an exponential run away in intelligence.

\subsection{The ``Fast Thinking Dog'' argument}

One of the arguments put forwards by proponents of the
technological singularity is that silicon has a significiant
speed advantage over our brain's wetware, and 
this advantage doubles every two years or so 
according to Moore's Law. 
Unfortunately speed along does 
not bring increased intelligence. To adapt an idea
from Vernor Vinge \cite{vinge}, 
a faster thinking dog is still unlikely to play chess.
Steven Pinker put this argument eloquently:
\begin{quote}
``{\em There is not the slightest reason to believe in a coming singularity. The fact that you can visualize a future in your imagination is not evidence that it is likely or even possible. Look at domed cities, jet-pack commuting, underwater cities, mile-high buildings, and nuclear-powered automobiles—all staples of futuristic fantasies when I was a child that have never arrived. Sheer processing power is not a pixie dust that magically solves all your problems.}'' \cite{pinker} 
\end{quote}
Intelligence is much more than thinking faster or
longer about a problem than someone else. 
Of course, Moore's Law has certainly helped AI. We now
learn off bigger data sets. We now learn quicker. 
Faster computers will certainly help us build
artificial intelligence. 
But, at least for humans, intelligence depends
on many other things including 
many years of experience and training. It is not
at all clear that we can short circuit this in
silicon simply by increasing the clock speed. 

\subsection{The ``Anthropcentric'' argument}

Many descriptions of the technological
singularity supposes human intelligence is some special
point to pass, some sort of ``tipping'' point. 
For instance, Nick Bostrom writes:
\begin{quote}
``{\em Human-level artificial intelligence leads quickly to greater-than-human-level artificial intelligence. \ldots  The interval during which the machines and humans are roughly matched will likely be brief. Shortly thereafter, humans will be unable to compete intellectually with artificial minds.}'' \cite{bostrom2}
\end{quote}
Human intelligence is 
one point on a wide spectrum that takes
us from cockroach through mouse to human. Actually, it might be 
better to say it is a probability distribution rather than a single
point. It is not clear in arguments like the above which level
of human intelligence requires to be exceeded before
run away growth kicks in. Is it some sort of average
intelligence? Or the intelligence of the smartest
human ever? 

If there is one thing that we should have
learnt from the history of science, it is that we are not
as special as we would like to believe. Copernicus taught us
that the universe did not revolve around the earth. Darwin
taught us that we were little different from the apes. 
And artificial intelligence will likely teach us
that human intelligence is itself nothing special. 
There is no reason therefore to suppose that human
intelligence is some special tipping point, that once passed
allows for rapid increases in intelligence. 
Of course, this doesn't preclude there being some level of intelligence which
is a tipping point. 

One argument put forwards by proponents of a technological
singularity is that human intelligence is indeed
a special point to pass because we are
unique in being able to build artefacts that amplify our 
intellectual abilities. We are the only creatures on the planet
with sufficient intelligence to design new intelligence, and
this new intelligence will not be limited by the slow process of
reproduction and evolution. 
However, this sort of argument supposes its conclusion. It 
assumes that human intelligence is enough to design an artificial
intelligence that is the sufficiently intelligent to be the
starting point for a technological singularity. In other words,
it assumes we have enough intelligence to initiate the
technological singularity, the very conclusion we are trying
to draw. We may or may not have enough intelligence
to be able to design such artificial intelligence. It is
far from inevitable. Even if have enough intelligence
to design super-human artificial
intelligence, this super-human artificial intelligence
may not be adequate to percipitate a
technological singularity. 

\subsection{The ``Meta-intelligence'' argument}

One of the strongest arguments against the
idea of a technological singularity in my view
is that it confuses intelligence to do a task with the
capability to improve your intelligence to do a task. 
David Chalmers, in an otherwise careful analysis of the 
idea of a technological singularity, 
writes:
\begin{quote}
{\em ``If we produce an AI by machine learning, it is likely that soon after we will be able to improve
the learning algorithm and extend the learning process, leading to AI+''} \cite{chalmers}
\end{quote} 
Here, AI is a system with human level intelligence and AI+ is 
a system more intelligent than the most intelligent human. 
But why should it be likely that soon after we can  improve 
the learning algorithm? Progress in machine learning algorithms
has neither been especially rapid or easy. 
Machine learning is indeed likely to be a significant
component of any human level AI system that we might build
in the future if only because it will be painful to
hand code its knowledge and expertise otherwise. 
Suppose an AI system uses machine learning
to improve its performance at some tasks requiring intelligence 
like understanding a text, or proving mathematical
identities. There is no reason that the system
can in addition improve the fundamental machine learning
algorithm used to do this. Machine learning algorithms
frequently top out a particular task, 
and no amount of tweaking, 
be it feature engineering or parameter tuning, 
appears able to improve their performance.

We are currently seeing impressive 
advances in AI using deep learning \cite{cacmdeeplearning}. 
This has dramatically improved the state-of-the-art in speech recognition, computer 
vision, natural language processing and a number of other domains. 
These improvements have come largely from using larger
data sets, and deeper neural networks:
\begin{quote}
{\em ``"Before, neural networks were not breaking records for
  recognizing continuous speech; they were not big enough."} 
Yann LeCun, quoted in \cite{cacmdeeplearning}
\end{quote}
Of coures, more data and bigger neural
networks means we need more processing power.
As a result, GPUs are now frequently used to provide
this processing power. 
However, being better able to recognize speech
or identify objects has not 
lead to an improvement in deep learning itself. 
The deep learning algorithms have not 
improved themselves. Any improvements to the
deep learning algorithms have
been hard won by applying our own intelligence
to their design. 

We can come at this argument from another direction
using one of the best examples we know of intelligent
systems. 
Look at ourselves. We only use a fraction of the
capabilities of our amazing brains, and we struggle to change this. 
It is much easier for us to learn how to do better at
a particular task, than it is for us to learn how to learn better
in general. For instance, if we remove the normalization inherent in the
definition of IQ, we can observe that 
IQ has increased over the last century but only slowly (the
``Flynn effect''). 
And improving your IQ today is pretty much as slow and painful
as it was a century ago. Perhaps electronic
brains will also struggle to improve their performance 
quickly and never get beyond a fraction of their 
fundamental capabilities?

\subsection{The ``Diminishing returns'' argument}

The idea of a technological singularity 
typically supposes improvements to intelligence will be a relative
constant multiplier, each generation getting some fraction better
that the last. However, 
the performance so far of most of our AI systems 
has been that of diminishing returns. 
There is often 
lots of low hanging fruit at the start, but 
we then run into great difficulties
to improve after this. This helps explain the overly
optimistic claims made by many of the early AI researchers. 
An AI system may be able to
improve itself an infinite number of
times, but the extent to which its intelligence changes
overall could be bounded. For instance, if 
each generation only improves by half the
last change, then the system will never get beyond
doubling its overall intelligence. 

Diminishing returns may also come not from 
the difficulty of improving our AI algorithms, but
from the difficulty of their subject matter increasing rapidly. 
Paul Allen, the Microsoft co-founder calls
this the ``complexity brake''. 
\begin{quote}
{\em ``We call this issue the complexity brake. As we go deeper and
  deeper in our understanding of natural systems, we typically find
  that we require more and more specialized knowledge to characterize
  them, and we are forced to continuously expand our scientific
  theories in more and more complex ways \ldots
we believe that progress toward this understanding [of cognition]
is fundamentally slowed by the complexity brake."} \cite{paulallen}
\end{quote}
Even if we see continual, perhaps even exponential improvements
in our AI systems, this may not be enough to improve
performance. The difficulty of the problems required
to be solved to see intelligence increase
may themselves increase even more rapidly. 
There are those that argue theoretical physics
appears to be running into such complexity brakes. 

\subsection{The ``Limits of intelligence'' argument}
 
There are many fundamental limits within the universe. 
Some of these are physical. 
You cannot accelerate past the speed of light. 
You cannot know both position and momentum with complete
accuracy. You cannot know when the radioactive decay of an atom
will happen with certainty. 
Any thinking machine that we might build will be limited by
these physical laws.
Of course, if that machine is electronic or even quantum in
nature, these limits are likely to be much greater than 
the biological and chemical limits of our human brains.

There are also more empirical laws which can be observed emerging
out of complex systems. For example, Dunbar's number
is the observed correlation between brain size for primates
and average social group size. This puts a limit of between
100 and 250 stable relationships on human social groups.
Intelligence is also a complex phenomenon and may also have 
such limits which emerge from this complexity. 
Any improvements in machine intelligence, 
whether it runs away or happens more slowly,
may run into such limits.
Or course, there is no reason to suppose that our own
human intelligence is at or close to this limit. But
equally, there's little reason why
any such limits are necessarily far beyond
our own intelligence. 

\subsection{The ``Computational complexity'' argument}

Suppose we stick to building AI systems with
computers that obey our traditional models
of computation. Even exponential improvements
are no match for computational complexity. 
For instance, exponential growth in performance is
inadequate to run super-exponential 
algorithms. And no amount of growth in performance
will make undecidable problems decidable.
Computational complexity may be one of the fundamental
limits discussed in the previous argument. 
Hence, unless we use machines that
beyond our traditional models of computation,
we are likely to bump into many 
problems where computational
complexity fundamentally limits
performance. 
Of course, a lot of computational complexity is
about worst case, and much of AI is about using heuristics
to solve problems in practice that are
computationally intractable in the worst
case. There are, however, fundamental
limits to the quality of these heuristics. 
There will be classes of problems that 
even a super-human intelligence cannot 
solve well, even approximately.

\section{Conclusions}

I have argued that there are many reasons
why we might never witness a technological
signularity. Nevertheless, even without a technological
singularity, we might still
end up with machines that exhibit super-human
levels of intelligence. We might just have
to program much of this painfully ourselves. 
If this is the case, 
the impact of AI on our economy, and on our society
may be less dramatic than either the pessimists
or the optimists have predicted. 
Nevertheless, we should start planning
for the impact that AI will have on society.
Even without a technological singularity,
AI is likely to have a large impact on the
nature of work. As a second example, even 
quite limited AI is likely to have a large impact
on the nature of war. We need to start planning
today for this future.

\bibliographystyle{aaai}
\bibliography{/Users/twalsh/Documents/biblio/a-z,/Users/twalsh/Documents/biblio/a-z2,/Users/twalsh/Documents/biblio/pub,/Users/twalsh/Documents/biblio/pub2}

\begin{thebibliography}{}

\bibitem[\protect\citeauthoryear{Allen and Greaves}{2011}]{paulallen}
Allen, P., and Greaves, M.
\newblock 2011.
\newblock The singularity isn't near.
\newblock {\em MIT Technology Review}  7--65.

\bibitem[\protect\citeauthoryear{Bostrom}{2002}]{bostrom2}
Bostrom, N.
\newblock 2002.
\newblock When machines outsmart humans.
\newblock {\em Futures} 35(7):759--764.

\bibitem[\protect\citeauthoryear{Bostrom}{2014}]{bostrom}
Bostrom, N.
\newblock 2014.
\newblock {\em Superintelligence: Paths, Dangers, Strategies}.
\newblock Oxford, UK: Oxford University Press.

\bibitem[\protect\citeauthoryear{Chalmers}{2010}]{chalmers}
Chalmers, D.
\newblock 2010.
\newblock The singularity: A philosophical analysis.
\newblock {\em Journal of Consciousness Studies} 17(9-10):7--65.

\bibitem[\protect\citeauthoryear{Edwards}{2015}]{cacmdeeplearning}
Edwards, C.
\newblock 2015.
\newblock Growing pains for deep learning.
\newblock {\em Commun. ACM} 58(7):14--16.

\bibitem[\protect\citeauthoryear{Good}{1965}]{good}
Good, I.
\newblock 1965.
\newblock Speculations concerning the first ultraintelligent machine.
\newblock {\em Advances in Computers} 6:31--88.

\bibitem[\protect\citeauthoryear{Kurzweil}{2006}]{kurzweil}
Kurzweil, R.
\newblock 2006.
\newblock {\em The Singularity Is Near: When Humans Transcend Biology}.
\newblock Penguin (Non-Classics).

\bibitem[\protect\citeauthoryear{Pinker}{2008}]{pinker}
Pinker, S.
\newblock 2008.
\newblock Tech luminaries address singularity.
\newblock {\em IEEE Spectrum}.

\bibitem[\protect\citeauthoryear{Turing}{1963}]{turing2}
Turing, A.
\newblock 1963.
\newblock Computing machinery and intelligence.
\newblock In {\em Computers and Thought}. McGraw-Hill Book Company.

\bibitem[\protect\citeauthoryear{Ulam}{1958}]{ulam}
Ulam, S.
\newblock 1958.
\newblock Tribute to {John} von {Neumann}.
\newblock {\em Bulletin of the American Mathematical Society} 64(3).

\bibitem[\protect\citeauthoryear{Vinge}{1993}]{vinge}
Vinge, V.
\newblock 1993.
\newblock The coming technological singularity: How to survive in the
  post-human era.
\newblock In Rheingold, H., ed., {\em Whole Earth Review}.

\end{thebibliography}

\end{document}